\pdfoutput=1

\documentclass{article}

\usepackage{arxiv}
\usepackage{xcolor}
\usepackage{graphicx}
\usepackage{amsmath}

\newcommand{\dname}{ASAD}

\usepackage[utf8]{inputenc} 
\usepackage[T1]{fontenc}    
\usepackage{hyperref}       
\usepackage{url}            
\usepackage{booktabs}       
\usepackage{amsfonts}       
\usepackage{nicefrac}       
\usepackage{microtype}      
\usepackage{lipsum}
\usepackage[misc,geometry]{ifsym} 

\title{\dname: A Twitter-based Benchmark  \underline{A}rabic \underline{S}entiment \underline{A}nalysis \underline{D}ataset}

\author{
   Basma Alharbi\thanks{The first three authors had equal contributions.}\\
   University of Jeddah \\
   Saudi Arabia \\
   \And
   Hind Alamro$^*$  \\
   KAUST \\
   Saudi Arabia \\
   \And
   Manal Alshehri$^*$  \\
   KAUST \\
  Saudi Arabia \\
   \And
   Zuhair Khayyat \\
   Lucidya \\
  Saudi Arabia \\
   \And
   Manal Kalkatawi \\
   KAU \\
  Saudi Arabia\\
  \And
  Inji Ibrahim Jaber\\
  KAUST\\
  Saudi Arabia\\
   \And
   Xiangliang Zhang$^{\textrm{\Letter}}$\thanks{$^{\textrm{\Letter}}$The corresponding author: Xiangliang Zhang,  Computer, Electrical and Mathematical Sciences and Engineering (CEMSE) Division and Computational Biosciences Research Center (CBRC), King Abdullah University of Science and Technology (KAUST). Email: xiangliang.zhang@kaust.edu.sa }\\
   KAUST \\
Saudi Arabia \\
}
\date{}

\begin{document}

\maketitle

\begin{abstract}
This paper provides a detailed description of a new Twitter-based benchmark dataset for Arabic Sentiment Analysis (\dname), which is launched in a  competition\footnote{\url{https://wti.kaust.edu.sa/arabic-sentiment-analysis-challenge}}, sponsored by KAUST for awarding 10000 USD, 5000 USD and 2000 USD to the first, second and third place winners, respectively.  Compared to other publicly released Arabic datasets, \dname\ is a large, high-quality annotated dataset (including 95K tweets), with three-class sentiment labels (\emph{positive, negative} and \emph{neutral}). We presents the details of the data collection process and annotation process. In addition, we implement several baseline models for the competition task and report the results as a reference for the participants to the competition. 
\end{abstract}


\section{Introduction} \label{sec:intro}

Sentiment analysis is a popular and challenging Natural Language Processing (NLP) task, which facilitates capturing public opinions on a topic, product or a service. It is thus a widely studied research area, where many benchmark datasets and algorithms exist to tackle this challenge. The state of research in Arabic sentiment analysis falls behind other high-source languages such as English and Chinese \cite{guellil2019arabic}. 
A comprehensive review on the current state-of-research on Arabic sentiment analysis can be found in \cite{al2019comprehensive}. In general, the work on Arabic sentiment analysis can be grouped by the employed approaches, which 
range from simple, rule-based / lexicon-based algorithms \cite{farra2010sentence} to more complex deep learning algorithms \cite{lecun2015deep}. Synthesizing recent publications on Arabic sentiment analysis show that there is an increase in the utilization of deep learning approaches in the past few years \cite{al2015deep,al2017hybrid,heikal2018sentiment,al2018deep,al2018deepsurvey,mohammed2019deep}. In such approaches, often a combination of word embedding model for feature representation is used along with a deep learning model for classification. For example, in
\cite{mohammed2019deep}, the authors applied three deep learning models: CNN, LSTM and RCNN, and the input to all these models were word embeddings generated from a pretrained word2vec CBOW model \cite{mikolov2013efficient}, named Aravec \cite{soliman2017aravec}. The contributions of work in Arabic deep learning-based sentiment analysis is limited to the application of existing models to Arabic corpora, and only a few publications contribute new models that are specifically designed to target the unique characteristics of the Arabic language, as in \cite{eljundi2019hulmona}.

The lack of large and high-quality Arabic benchmark dataset is often attributed to this deficiency in Arabic-based NLP studies \cite{nabil2014labr}. The objective of this work is to introduce a new large Twitter-based Arabic Sentiment Analysis Dataset (\dname). \dname\ is a public dataset intended to accelerate research in Arabic NLP in general, and Arabic sentiment classification in specific. 
A  competition\footnote{Competition website: \url{https://wti.kaust.edu.sa/arabic-sentiment-analysis-challenge}}  is launched on top of the \dname\ dataset.





This being said, the increasing interest in the past few years in Arabic sentiment analysis has resulted in the release of a number of datasets. Datasets collected for Arabic sentiment analysis span multiple sources, which include newspapers, reviews, and different content from social media platforms \cite{abdul2014sana, aly2013labr, elsahar2015building, abdul2012awatif}. Table \ref{tab:rw} compares our dataset \dname\ with similar benchmark datasets, which are also Twitter-based Arabic   corpora for general sentiments (as opposed to topic-based sentiment). 
The benchmark datasets are compared in terms of the size of dataset (i.e., the number of annotated tweets), the number of sentiment classes,  the inclusion of Arabic dialect, the annotation approach, the number of annotators and the quality of the annotation (measured by Fleiss's Kappa metric \cite{fleiss1971measuring}). 

TEAD \cite{abdellaoui2018using}, the currently largest dataset for Arabic sentiment analysis, was annotated automatically using emojis in tweets. The same automatic annotation approach was adopted by ASTAD \cite{kwaik2020arabic}, while no explicit details of how the 40k Tweets dataset was annotated \cite{mohammed2019deep}. The remaining datasets have significantly smaller size, yet they were all annotated manually. Manual annotations indicate that a number of human annotators were recruited and trained to provide high quality annotations for each record in the dataset. Manual annotation thus provides a more reliable data source for sentiment classification algorithms to learn from. The quality of annotations can be assessed, e.g.,   
by the most commomly used Fleiss's Kappa metric \cite{fleiss1971measuring}, whose values closer to 1 indicate higher agreements between annotators. Among the four manually annotated corpora, AraSenTi-Tweet \cite{al2017arasenti} and Gold Standard \cite{refaee2014arabic} stated their annotation quality measure, which are included in Table 1. 

SemEval \cite{rosenthal2017semeval} annotates tweets in either one of five sentiments, which are: \emph{strongly positive, weakly positive, neutral, weakly negative} and \emph{strongly negative}. AraSenTi-Tweet \cite{al2017arasenti}, ASTD \cite{nabil2015astd}, and Gold standard \cite{refaee2014arabic} adopt a four-point scale for sentiments, which are: \emph{positive, neutral, negative} and \emph{mixed classes}. Lastly, TEAD \cite{abdellaoui2018using} follows a three-point scale for sentiments, where each tweet is either \emph{positive, negative} or \emph{neutral}. Both 40K Tweets \cite{mohammed2019deep} and ASTAD \cite{kwaik2020arabic} use a two-point scale for sentiment, where tweets are either positive or negative. 
Our dataset, \dname, adopts a three-point scale similarly to TEAD \cite{abdellaoui2018using}, has multiple dialects, and was annotated manually by an average of three annotators per tweet. The annotation quality for our dataset is $\kappa = 0.56$, indicating a moderate agreement between the annotators.  A detailed analysis of the annotation is presented in the next section. 

The rest of this paper is organized as follows. Section \ref{sec:data} provides details on the construction mechanism we followed to collect \dname. Section \ref{sec:eval} provides details on becnhmark evaluation and results using baseline approaches for three-class sentiment classification. Section \ref{sec:dis} provides a discussion on challenges and future applications to \dname. 



\begin{table}[t!]
\caption{Summary of  Twitter-based Arabic Sentiment Analysis Corpora. The '-' sign indicates that these metrics were not explicitly specified in the paper.}
\label{tab:rw}
\centering
\small
\begin{tabular}{|p{80pt}|p{30pt}|p{28pt}|p{45pt}|p{50pt}|p{45pt}| p{45pt}|}
\hline
                         & \textbf{Size} & \textbf{No. Classes} & \textbf{Dialect}  & \textbf{Annotation \newline Approach} & \textbf{No. of \newline Annotators}& \textbf{Annotation Quality}      \\ \hline
\textbf{TEAD} \cite{abdellaoui2018using}  &      5,500 M  & 3 &    Multi    &                           from Emojis & - & -      \\ \hline
\textbf{40K Tweets} \cite{mohammed2019deep}  &      40,000  & 2 &   MSA, Egyptian    &                           - & - & -      \\ \hline
\textbf{ASTAD} \cite{kwaik2020arabic}  &      36,000  & 2 &   Multi    &                           from Emojis & - & -      \\ \hline
\textbf{Ara-SenTi-Tweet} \cite{al2017arasenti} &   17,573                         & 4 & MSA, Saudi       & Manual & 3 &$\kappa = 0.60$\\ \hline
\textbf{ASTD} \cite{nabil2015astd}                  &  10,000                         & 4 & Egyptian         & Manual & 3 & - \\ \hline
\textbf{Gold   Standard} \cite{refaee2014arabic}  & 8,868 & 4 & Multi & Manual & 2 & $\kappa = 0.82$ \\ \hline
\textbf{SemEval} \cite{rosenthal2017semeval}   &        9,455   & 5 &  NA   & Manual & 5 &     -                          \\ \hline
Our \textbf{\dname}      &   100,000   & 3 &    Multi &  Manual & 3 &   $\kappa = 0.56$  \\ \hline

\end{tabular}

\end{table}

\section{Corpus Construction} \label{sec:data}

\begin{figure}
	\centering
		\includegraphics[width=0.8\textwidth ]{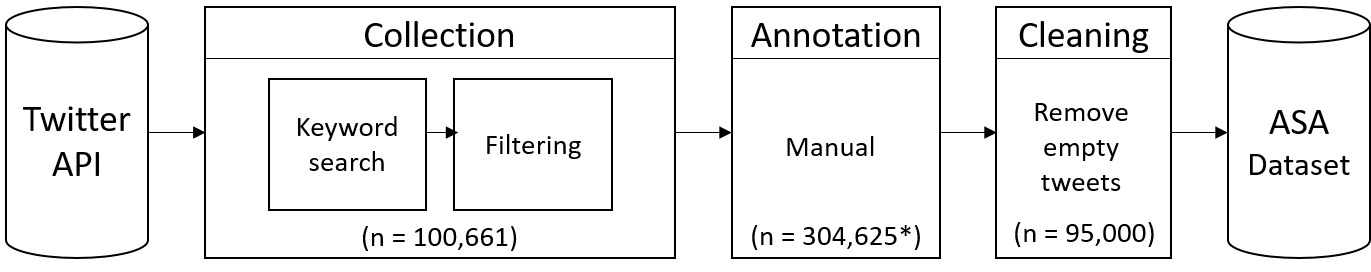}
	\caption{Corpus Collection Methodology. Illustration of the different phases involved in collecting, annotating and cleaning the dataset. * In the annotation phase, each record was annotated by at least three annotators. This justifies the increase in the number of records in the annotation phase when compare to the previous phase (the data collection phase).}
	\label{fig:corps_collection}
\end{figure}

Figure \ref{fig:corps_collection} illustrates the methodology we adopted to construct the corpus. In general, the methodology consists of three main phases: data collection, data annotation and data cleaning. 
Each phase is further described in the subsections below. The data collection and annotation was done by Lucidya\footnote{\url{https://lucidya.com/}} which is an AI-based company with rich experience in organizing data annotation projects.

\subsection{Corpus Collection}


The tweets in {\dname} were randomly selected from a pool of tweets that have been collected using the Twitter public streaming API\footnote{\url{https://developer.twitter.com/en/docs/twitter-api}} between May 2012 and April 2020. The selection process consisted of several filtration operations to ensure that the Arabic texts are comprehensive enough for the annotation task. First, a random set of tweets with "ar" label in their json object were selected from the pool of tweets and added to the corpus. Next, tweets that contained less than seven words (excluding hashtags and user mentions) were removed from the corpus. After that, all tweets that included URLs, images, or videos or consisted of a predefined set of inappropriate Arabic keywords were also removed from the Corpus. Finally, the corpus was refined to include 100k tweets. 

We analyzed the content of {\dname} and found out that around 69\% of the tweets were selected from the year 2020 (January 2020 until April 2020), 30\% were tweeted in 2019 and the remaining 1\% happened between 2012 and 2018. 82\% of the tweets came from unique authors where we encountered only 503 verified authors in the corpus. For locations, 72\% of the tweets originated from Saudi Arabia, 13\% from Egypt, 7\% from Kuwait, 3\% from the United States, and 3\% from the United Arab Emirates. Based on the models developed by Lucidya, 
36\% of the tweets are written in Modern Arabic, 31\% of the tweets used the Khaleeji dialect, 22\% used the Hijazi dialect, and the rest 10\% tweeted in Egyptian dialect.

\subsection{Corpus Annotation}
\begin{figure}
	\centering
		\includegraphics[width=0.35\textwidth ]{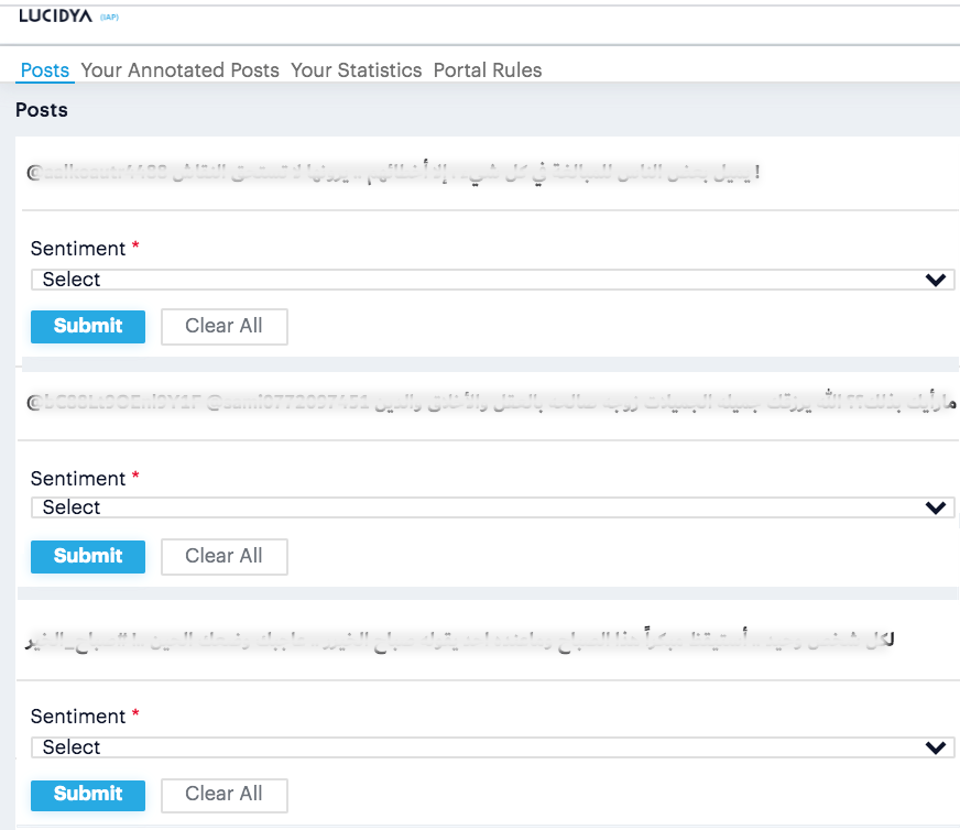}
	\caption{Screenshot of the Annotation Platform.}
	\label{fig:corps_annotation}
\end{figure}

The objective of this phase is to apply tweet-level sentiment annotation. We adopted a three-way classification sentiment, where a sentiment can either be: \emph{positive, negative} or \emph{neutral}. 
A screenshot of the platform for our project is given in Figure \ref{fig:corps_annotation}. Each tweet was annotated by at least three experienced annotators who were trained and qualified by the company.

 
The annotation process was done by 69 native Arabic speakers with an average age of 26 years old where 12\% of the recruited annotators were under the age of 21, 71\% between 21 and 30, 13\% between 31 and 40, and the rest over 40 years old. Among the annotators, 85\% of them were females, and 71\% of them  have at least a Bachelor's degree, and 4\% obtained a degree in higher education. In addition, 67\% of the annotators have a background in arts, linguistics, or education while the remaining 33\% majored in either science or engineering. Based on Lucidya's experience, it is more efficient to hire local annotators which resulted in having 71\% of the annotators to be from Saudi Arabia. The remaining annotators came from various parts of the Middle East where 12\% were hired from Egypt, 7\% from Yemen, 7\% from Algeria, while the rest came from Indonesian and Palestine.
 
The annotators started the corpus annotation project on February 2020 and finished by May 2020 where the tweets were added to the the annotation portal in several batches\footnote{Each batch of size $10,000$ tweets.}. By the end of each batch, an evaluation of the quality of the annotation was conducted to ensure that only high quality annotators were allowed to participate in the next batch. By the end of the 10 batches, we gained a total of $304,625$ annotations for $100,661$ distinct tweets, with a ratio of $~ 3$ annotations per tweet. 
\subsection{Data Cleaning} 
Preliminary data cleaning was done to ensure minimum data quality level. Empty tweets, i.e., tweets with no text, as well as meaningless tweets (e.g., those containing only URLs or coupons) were excluded from the corpus. The total number of identified records at this stage is 95,000.

\subsubsection{Annotation Evaluation} \label{sec:ann_eval}
For a corpus to be considered as a benchmark, we need to verify the reliability of the labels as well as the difficulty level of the task. For that, we adopt the 
Fleiss’s  Kappa  \cite{fleiss1971measuring}  metric to measure the  Inter  Annotator  Agreement (IAA). 
The Fleiss's Kappa measure is calculated by
\begin{align} \small
   \kappa &= \frac{\bar{P} - \bar{P}_e}{1 - \bar{P}_e} ; \; \text{where}  \\
   \bar{P} & = \frac{1}{N} \sum_{i=1}^N  \frac{1}{n(n-1)} \sum_{j=1}^k n_{ij} (n_{ij} - 1) \\ 
   \bar{P}_e &= \sum_{j=1}^k ( \frac{1}{Nn} \sum_{i=1}^N n_{ij})^2
\end{align} 
where $N$ is the total number of tweets, $n$ is the number of annotations per tweet, $k$ is the number of classes, and $n_{ij}$ is the number of annotators who have assigned the $i$-th tweet to the $j$-th class. 
{  In our \dname\ dataset, we have tweets annotated by three annotators (n=3), and by four annotators (n=4). The agreement among the annotators and the calculated $\kappa$ for each case is shown in Table \ref{tab:kappa}. 
The over all Fleiss Kappa coefficient of  \dname\ is $\kappa$=$0.56$, indicating a moderate agreement among the annotators.}

\begin{table}[t]
\caption{The Fleiss Kappa coefficient $\kappa$ of \dname}
\centering
\begin{tabular}{|l|l|l|l|l|}
\hline
No. annotators  & \multicolumn{3}{c|}{No. tweets ($N$)} & $\kappa$ \\ \cline{2-4}
per tweet  & with 1 agreed label & with 2 different labels & with 3 different labels & \\
\hline
$n$=3 & 58424 (62.34\%) & 35285 (37.65\%) &10 (0.01\%) & 0.5035\\
\hline
$n$=4 & 813 (63.47\%) & 439(34.27\%) & 29(2.26\%) & 0.6067 \\
\hline
\multicolumn{5}{|r|}{averaged $\kappa$=  0.56} \\ \hline
\end{tabular}
\label{tab:kappa}
\end{table}

\begin{figure}
	\centering
		\includegraphics[width=0.48\textwidth ]{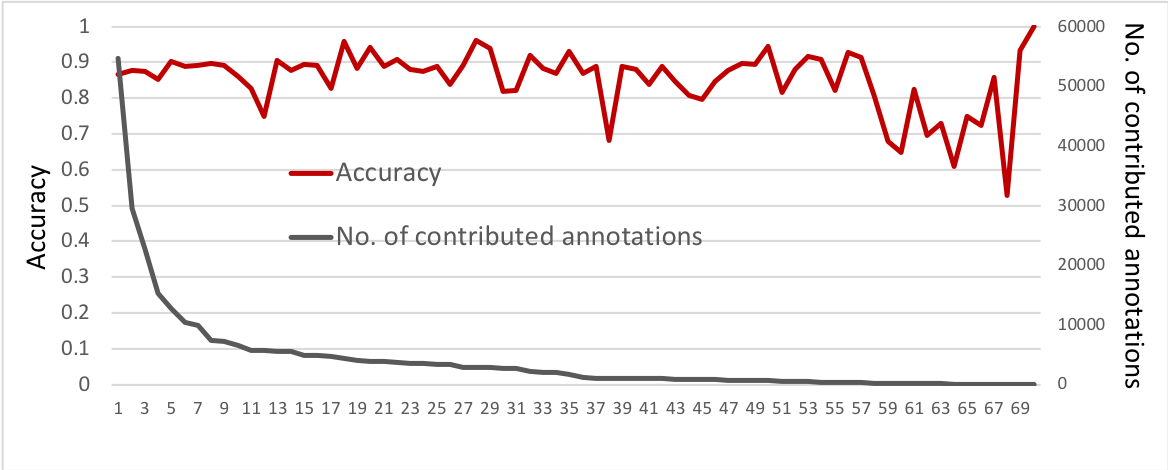}
	\caption{The accuracy of  69 annotators, ordered by their number of contributed annotations.}
	\label{fig:annotator}
\end{figure}

{ Since there are around 37\% tweets initially annotated with more than two different labels (as given in Table \ref{tab:kappa}), we investigate whether the disagreement is due to the fact that the labeling task is  non-trivial for human annotators, or is caused by the carelessness of certain annotators. We determine the final label for each tweet   by majority voting or another experienced annotator if there was no consensus. Then, we evaluate the accuracy of each annotator: how much percentage of their annotated tweets are the same as the final label. The accuracy is shown in Figure \ref{fig:annotator}, where the 69 annotators are ordered by their number of contributed annotations. We can see that the top-10 annotators contributed more than 60\% annotations with an average accuracy 0.88. The overall  average accuracy is 0.85 regardless the number of contributed annotations. Only a few annotators had accuracy less than 0.8, and most of these annotators gave less than 200 annotations. Following  \cite{SemEval2018Task1} and \cite{bobicev2017inter}, { we also evaluate the average inter-rater agreement $\iota$= $0.83$, indicating the reliability of the annotators   since  $\iota$ is much higher than $0.5$}.  Therefore, we can conclude that the annotators had conducted high-quality labeling work. The disagreement is mainly caused by the non-travail understanding of the tweets themselves. Such challenges in real-world tweets sentiment annotation also make our competition extremely meaningful. We expect the advanced models developed by the participants can be the effective solutions to this problem.  }

\subsection{Corpus Properties} 
\dname\ consists of 95,000 annotated tweets, with three-class sentiments. 
Table \ref{tab:datasplits} presents basic statistics about \dname. \dname\ (represented by ALL in the table) has a total of 15,215 \emph{positive} tweets, 15,267 \emph{negative} tweets and 64,518 \emph{neutral} tweets. 
In the competition, the entire \dname\ data is split into \emph{TRAINING, TEST1} and \emph{TEST2} for the usage at different stages. The data distribution in all subset splits are the same. We also use these data split for benchmark evaluation, which is described in Section \ref{sec:eval}. 


\section{Benchmark Evaluation} \label{sec:eval}


This section provides details on evaluation details and baseline results. 

\subsection{Competition and Evaluation Details}

In the competition,  the \dname\ dataset is divided into three folds: training (TRAINING), development-time testing (TEST1) and main testing (TEST2). The TRAINING set contains 55K labeled tweets, while TEST1 and TEST2 are composed of 20K tweets each, as shown in Table \ref{tab:datasplits}. 
In the competition, both TRAINING and TEST1 datasets are released at the beginning  of the competition. At this stage, competing teams can assess their relative rank using the evaluation metric described next on the TEST1 set.  At the end of the competition, teams will be ranked based on their performance on TEST2. 
Teams will be ranked on TEST2, where ranking will be done offline by evaluating the submitted codes. The prizes for the first three winners are: 
\begin{itemize}
\item First place winner, 10,000 USD 
\item Second place winner, 5,000 USD 
\item Third place winner, 2,000 USD 
\end{itemize}

At the end of the competition, a report will be published by inviting the top ranking teams to describe their winning solutions. And an awarding conference  will be organized to invite the team introducing their solutions.
For more details about the competition, timelines and rules please visit the competition website: \url{https://wti.kaust.edu.sa/arabic-sentiment-analysis-challenge}. 

\begin{table}
\caption{Class distribution in data splits.
The \emph{All} set represents the complete collected dataset. \emph{TRAINING, TEST1} and \emph{TEST2} are the three splits used in the competition and also in the benchmark models evaluation. The three splits maintained the same class distribution as the original complete dataset.}
\centering
\begin{tabular}{|l|l|l|l|l|l|l|l|l|}
\hline
  & \multicolumn{2}{c|}{\textbf{TRAINING}} & \multicolumn{2}{c|}{\textbf{TEST1}} & \multicolumn{2}{c|}{\textbf{TEST2}} & \multicolumn{2}{c|}{\textbf{All}} \\ \cline{2-9} 
                  & No. tweets &   (\%) & No. tweets &     (\%) & No. tweets &     (\%) & No. tweets &   (\%) \\ \hline
\textbf{Positive} & 8821  & 0.16          & 3150  & 0.16          & 3244  & 0.16          & 15215 & 0.16          \\ \hline
\textbf{Negative} & 8820  & 0.16          & 3252  & 0.16          & 3195  & 0.16          & 15267 & 0.16          \\ \hline
\textbf{Neutral}  & 37359 & 0.68          & 13598 & 0.68          & 13561 & 0.68          & 64518 & 0.68          \\ \hline
\textbf{Total}    & 55000 & 1.00          & 20000 & 1.00          & 20000 & 1.00          & 95000 & 1.00          \\ \hline
\end{tabular}

\label{tab:datasplits}
\end{table}

In addition to the popular Macro-F1 score and Micro-F1 score, we use the following metrics for evaluating the performance of various models on this task. 
\begin{itemize}
\item \textbf{Primary metric},  the macro-averaged recall (recall averaged across the three classes)
\begin{align} 
AvgRec &= \frac{1}{3} (R^P + R^N + R^U)  
\end{align}
where $R^P, R^N$ and $R^U$ are the recall for the positive, negative and neutral class, respectively. 
\item \textbf{Secondary metric}, the average $F_1$ for positive and negative classes only
\begin{align} 
F_1^{PN} &= \frac{1}{2} (F_1^P + F_1^N) 
\end{align}
where $F_1^P$ and $F_1^N$ refer to the $F1$ measure for the positive and negative classes, respectively. For more details on the evaluation metric, please refer to \cite{nakov2019semeval}. 
\end{itemize}

In the competition, participants will be evaluated and ranked by their performance on the  metric of {\bf Marco-F1 score}. 

\subsection{Our Baselines and Results} 
\textbf{Preprocessing:}
In our baseline experiments we applied some preprocessing to the data by removing the stop-words, special symbols like ‘@’ and ‘\#’, and URLs. We also removed non-Arabic characters in order to make a clear analysis of Arabic language. In addition, the word tokenization was applied to the text data. For this part we used the NLTK (https://www.nltk.org )  and pyarabic tool (https://github.com/linuxscout/pyarabic). 

\textbf{Traditional Baseline Methods:} We applied bag-of-words (BOW) and Tf-Idf to extract the features of the text data, then for each of them we classify by using logistic regression.

\begin{itemize}
\item \textbf{BOW+ Logistic regression.}
The BOW is the simplest form of text representation, which describes the occurrence of words within a document. We used BOW to extract the features of the texts, then we fed them into the logistic regression classifier with 'lbfgs' solver.
\item\textbf{TF-IDF + Logistic regression.}
TF-IDF  measures how important a particular word is with respect to a document and the entire corpus. Similar to the previous model, the text features where extracted by Tf-Idf, then we applied the logistic regression classifier. 
\end{itemize} 

For BOW, TF-Idf, and logistic regression, we used the popular Scikit-learn\footnote{\url{https://scikit-learn.org/stable/getting_started.html}} python library.
Scikit-learn is an open source machine learning library that supports supervised and unsupervised learning. It also provides various tools for model fitting, data preprocessing, model selection and evaluation, and many other utilities. 

\textbf{Deep Learning Baseline Models:} 
We fine-tuned the pre-trained model BERT \cite{devlin2018bert} and AraBERT \cite{antoun2020arabert}  for learning the embedding vectors for tweets and then for the classification. 
\begin{itemize}
\item \textbf{BERT} \cite{devlin2018bert}
is one of the most popular language representation models developed by Google.  
BERT can be fine-tuned with one additional output layer to create a model for a specific NLP task. Several multilingual BERT models have been released to support many languages.
\item \textbf{AraBERT} \cite{antoun2020arabert}
is an Arabic language model by training BERT on a large-scale Arabic corpus. It overcomes the performance of a multilingual version of BERT in many NLP tasks applied to Arabic text.
\end{itemize} 
To fine-tune BERT and AraBERT, we used ktrain\footnote{\url{https://github.com/amaiya/ktrain}}, which is a lightweight wrapper for many deep learning libraries to help build, train, and deploy machine learning models in a more accessible and easier manner. With ktrain, the text must be pre-processed in a specific way to obtain a fixed-size sequence of word IDs for each document. Then, they are used as input for the classifier. Each model in ktrain has its corresponding pre-processing procedure and classifier.

Table \ref{tab:baseline} presents  the  results of the baseline models when trained by TRAINING and evaluated on TEST1 and TEST2. These initial results provide a baseline for the competing teams, where a significant improvement is expected. Additionally, it is clear that the evaluation results on TEST1 is statistically similar to those obtained when testing the same model on TEST2. This indicates that the relative ranking on TEST1 is a good indicator for the participating teams of their final ranking on TEST2. Comparing the results of different models, we see AraBert has the best performance in terms of $Avg-Rec$ and $F_1^{PN}$, as well as the general metrics Macro-F1 and Micro-F1.

\begin{table}[t]
\caption{Evaluation results of baseline models on TEST1 and TEST 2}
\centering
\begin{tabular}{|l|l|l|l|l|l|l|l|l|}
\hline
      & \multicolumn{2}{l|}{BOW-LR} & \multicolumn{2}{l|}{TF-IDF-LR} & \multicolumn{2}{l|}{BERT} & \multicolumn{2}{l|}{AraBERT} \\ \cline{2-9} 
  & TEST1           & TEST2          & TEST1          & TEST2          & TEST1         & TEST2       & TEST1         & TEST2      \\ \hline
F1-Positive &0.51  &0.51 &0.52   &0.53   &0.62  &0.61  &0.65  &0.65 \\ \hline
F1-Negative &0.49  &0.48 &0.51    &0.50  &0.39  &0.38  &0.53  &0.52 \\ \hline
F1-Neutral &0.83  &0.83 &0.83    & 0.83  &0.85  &0.85  &0.86  &0.87 \\ \hline
Macro-F1 &0.61  &0.60 &0.62    & 0.62    &0.62  &0.62  &0.68  &0.68 \\ \hline
Micro-F1 &0.72  & 0.72 &0.73  & 0.73     &0.74  &0.74  &0.78  &0.78 \\ \hline
Positive-Recall &0.43  & 0.43 &0.47    & 0.47 &0.60  &0.59  &0.64  &0.63 \\ \hline
Negative-Recall &0.43  &0.42 &0.47   &0.45    &0.28  &0.27  &0.45  &0.43 \\ \hline
Neutral-Recall &0.88  &0.88 & 0.87  & 0.87   &0.92  &0.92  &0.90  &0.91 \\ \hline\hline
$Avg-Rec$ &0.58  &0.58  & 0.60 & 0.60 &0.60  &0.59 &{\bf 0.66 } &{\bf 0.66} \\ \hline
$F_1^{PN}$ &0.50  &0.50 &0.52  &0.52  &0.50  &0.50 &{\bf 0.59 } &{\bf 0.59} \\ \hline
\end{tabular}
\label{tab:baseline}
\end{table}

\section{Discussion} \label{sec:dis}

In this section, we highlight some of the challenges related to Arabic sentiment classification from a Twitter-based dataset. We identify three main challenges, listed below: 

\begin{itemize} 

\item \textbf{Sentiment Imbalance: } \dname\ adopts a three-point sentiment scale, where a tweet is labeled as either \emph{positive}, \emph{negative} or \emph{neutral}. As it was detailed in Section \ref{sec:data}, the distribution of the labels is imbalanced, where the \emph{neutral} class has more number of tweets than the other two classes. Imbalanced datasets create a challenge to most learning algorithms. Thus, the prediction performance of the model should be evaluated using a metric that handles unbalanced data.

\item \textbf{Spam: } The \dname\ dataset was preliminary cleaned as discussed in Section \ref{sec:data}. However, spam tweets still exist and may affect the prediction performance of a sentiment classifier. 

\item \textbf{Multiple Dialects: } The existence of multiple dialects in the dataset creates a challenge to the sentiment classification problem. In a simpler dataset with one dialect only, e.g., ASTD \cite{nabil2015astd}, a supervised learning algorithm will be able to identify the main keywords associated with each sentiment in this dialect. When multiple dialects exists, the range of keywords associated with each sentiment increases because such words differ from one dialect to another. This creates an additional challenge that one-dialect corpus does not have. 
A pre-trained word embedding model with one-dialect corpus may not work well on the multi-dialect dataset. 


\end{itemize} 

The complete dataset with both training and testing sets will be publicly released for the research community after the competition ends. The dataset can be used for a range of other NLP tasks, besides sentiment classification. This includes dialect identification, spam detection, and also assist the sentiment analysis of Covid-19 tweets \cite{yang2020senwave}. We will update our work in the near future. 




\bibliographystyle{unsrt}  

\bibliography{references} 

\begin{thebibliography}{10}

\bibitem{guellil2019arabic}
Imane Guellil, Houda Sa{\^a}dane, Faical Azouaou, Billel Gueni, and Damien
  Nouvel.
\newblock Arabic natural language processing: An overview.
\newblock {\em Journal of King Saud University-Computer and Information
  Sciences}, 2019.

\bibitem{al2019comprehensive}
Mahmoud Al-Ayyoub, Abed~Allah Khamaiseh, Yaser Jararweh, and Mohammed~N
  Al-Kabi.
\newblock A comprehensive survey of arabic sentiment analysis.
\newblock {\em Information processing \& management}, 56(2):320--342, 2019.

\bibitem{farra2010sentence}
Noura Farra, Elie Challita, Rawad Abou~Assi, and Hazem Hajj.
\newblock Sentence-level and document-level sentiment mining for arabic texts.
\newblock In {\em 2010 IEEE international conference on data mining workshops},
  pages 1114--1119. IEEE, 2010.

\bibitem{lecun2015deep}
Yann LeCun, Yoshua Bengio, and Geoffrey Hinton.
\newblock Deep learning.
\newblock {\em nature}, 521(7553):436--444, 2015.

\bibitem{al2015deep}
Ahmad Al~Sallab, Hazem Hajj, Gilbert Badaro, Ramy Baly, Wassim El-Hajj, and
  Khaled Shaban.
\newblock Deep learning models for sentiment analysis in arabic.
\newblock In {\em Proceedings of the second workshop on Arabic natural language
  processing}, pages 9--17, 2015.

\bibitem{al2017hybrid}
Sadam Al-Azani and El-Sayed~M El-Alfy.
\newblock Hybrid deep learning for sentiment polarity determination of arabic
  microblogs.
\newblock In {\em International Conference on Neural Information Processing},
  pages 491--500. Springer, 2017.

\bibitem{heikal2018sentiment}
Maha Heikal, Marwan Torki, and Nagwa El-Makky.
\newblock Sentiment analysis of arabic tweets using deep learning.
\newblock {\em Procedia Computer Science}, 142:114--122, 2018.

\bibitem{al2018deep}
Mohammad Al-Smadi, Omar Qawasmeh, Mahmoud Al-Ayyoub, Yaser Jararweh, and Brij
  Gupta.
\newblock Deep recurrent neural network vs. support vector machine for
  aspect-based sentiment analysis of arabic hotels’ reviews.
\newblock {\em Journal of computational science}, 27:386--393, 2018.

\bibitem{al2018deepsurvey}
Mahmoud Al-Ayyoub, Aya Nuseir, Kholoud Alsmearat, Yaser Jararweh, and Brij
  Gupta.
\newblock Deep learning for arabic nlp: A survey.
\newblock {\em Journal of computational science}, 26:522--531, 2018.

\bibitem{mohammed2019deep}
Ammar Mohammed and Rania Kora.
\newblock Deep learning approaches for arabic sentiment analysis.
\newblock {\em Social Network Analysis and Mining}, 9(1):52, 2019.

\bibitem{mikolov2013efficient}
Tomas Mikolov, Kai Chen, Greg Corrado, and Jeffrey Dean.
\newblock Efficient estimation of word representations in vector space.
\newblock {\em arXiv preprint arXiv:1301.3781}, 2013.

\bibitem{soliman2017aravec}
Abu~Bakr Soliman, Kareem Eissa, and Samhaa~R El-Beltagy.
\newblock Aravec: A set of arabic word embedding models for use in arabic nlp.
\newblock {\em Procedia Computer Science}, 117:256--265, 2017.

\bibitem{eljundi2019hulmona}
Obeida ElJundi, Wissam Antoun, Nour El~Droubi, Hazem Hajj, Wassim El-Hajj, and
  Khaled Shaban.
\newblock hulmona: The universal language model in arabic.
\newblock In {\em Proceedings of the Fourth Arabic Natural Language Processing
  Workshop}, pages 68--77, 2019.

\bibitem{nabil2014labr}
Mahmoud Nabil, Mohamed Aly, and Amir Atiya.
\newblock Labr: A large scale arabic sentiment analysis benchmark.
\newblock {\em arXiv preprint arXiv:1411.6718}, 2014.

\bibitem{abdul2014sana}
Muhammad Abdul-Mageed and Mona~T Diab.
\newblock Sana: A large scale multi-genre, multi-dialect lexicon for arabic
  subjectivity and sentiment analysis.
\newblock In {\em LREC}, pages 1162--1169, 2014.

\bibitem{aly2013labr}
Mohamed Aly and Amir Atiya.
\newblock Labr: A large scale arabic book reviews dataset.
\newblock In {\em Proceedings of the 51st Annual Meeting of the Association for
  Computational Linguistics (Volume 2: Short Papers)}, pages 494--498, 2013.

\bibitem{elsahar2015building}
Hady ElSahar and Samhaa~R El-Beltagy.
\newblock Building large arabic multi-domain resources for sentiment analysis.
\newblock In {\em International Conference on Intelligent Text Processing and
  Computational Linguistics}, pages 23--34. Springer, 2015.

\bibitem{abdul2012awatif}
Muhammad Abdul-Mageed and Mona~T Diab.
\newblock Awatif: A multi-genre corpus for modern standard arabic subjectivity
  and sentiment analysis.
\newblock In {\em LREC}, volume 515, pages 3907--3914, 2012.

\bibitem{fleiss1971measuring}
Joseph~L Fleiss.
\newblock Measuring nominal scale agreement among many raters.
\newblock {\em Psychological bulletin}, 76(5):378, 1971.

\bibitem{abdellaoui2018using}
Houssem Abdellaoui and Mounir Zrigui.
\newblock Using tweets and emojis to build tead: an arabic dataset for
  sentiment analysis.
\newblock {\em Computaci{\'o}n y Sistemas}, 22(3):777--786, 2018.

\bibitem{kwaik2020arabic}
Kathrein~Abu Kwaik, Stergios Chatzikyriakidis, Simon Dobnik, Motaz Saad, and
  Richard Johansson.
\newblock An arabic tweets sentiment analysis dataset (atsad) using distant
  supervision and self training.
\newblock In {\em Proceedings of the 4th Workshop on Open-Source Arabic Corpora
  and Processing Tools, with a Shared Task on Offensive Language Detection},
  pages 1--8, 2020.

\bibitem{al2017arasenti}
Nora Al-Twairesh, Hend Al-Khalifa, AbdulMalik Al-Salman, and Yousef Al-Ohali.
\newblock Arasenti-tweet: A corpus for arabic sentiment analysis of saudi
  tweets.
\newblock {\em Procedia Computer Science}, 117:63--72, 2017.

\bibitem{refaee2014arabic}
Eshrag Refaee and Verena Rieser.
\newblock An arabic twitter corpus for subjectivity and sentiment analysis.
\newblock In {\em LREC}, pages 2268--2273, 2014.

\bibitem{rosenthal2017semeval}
Sara Rosenthal, Noura Farra, and Preslav Nakov.
\newblock Semeval-2017 task 4: Sentiment analysis in twitter.
\newblock In {\em Proceedings of the 11th international workshop on semantic
  evaluation (SemEval-2017)}, pages 502--518, 2017.

\bibitem{nabil2015astd}
Mahmoud Nabil, Mohamed Aly, and Amir Atiya.
\newblock Astd: Arabic sentiment tweets dataset.
\newblock In {\em Proceedings of the 2015 conference on empirical methods in
  natural language processing}, pages 2515--2519, 2015.

\bibitem{SemEval2018Task1}
Saif~M. Mohammad, Felipe Bravo-Marquez, Mohammad Salameh, and Svetlana
  Kiritchenko.
\newblock Semeval-2018 {T}ask 1: {A}ffect in tweets.
\newblock In {\em Proceedings of International Workshop on Semantic Evaluation
  (SemEval-2018)}, New Orleans, LA, USA, 2018.

\bibitem{bobicev2017inter}
Victoria Bobicev and Marina Sokolova.
\newblock Inter-annotator agreement in sentiment analysis: Machine learning
  perspective.
\newblock In {\em RANLP}, pages 97--102, 2017.

\bibitem{sebastiani2015axiomatically}
Fabrizio Sebastiani.
\newblock An axiomatically derived measure for the evaluation of classification
  algorithms.
\newblock In {\em Proceedings of the 2015 International Conference on The
  Theory of Information Retrieval}, pages 11--20, 2015.

\bibitem{nakov2019semeval}
Preslav Nakov, Alan Ritter, Sara Rosenthal, Fabrizio Sebastiani, and Veselin
  Stoyanov.
\newblock Semeval-2016 task 4: Sentiment analysis in twitter.
\newblock {\em arXiv preprint arXiv:1912.01973}, 2019.

\bibitem{devlin2018bert}
Jacob Devlin, Ming-Wei Chang, Kenton Lee, and Kristina Toutanova.
\newblock Bert: Pre-training of deep bidirectional transformers for language
  understanding.
\newblock {\em arXiv preprint arXiv:1810.04805}, 2018.

\bibitem{antoun2020arabert}
Wissam Antoun, Fady Baly, and Hazem Hajj.
\newblock Arabert: Transformer-based model for arabic language understanding,
  2020.

\bibitem{yang2020senwave}
Qiang Yang, Hind Alamro, Somayah Albaradei, Adil Salhi, Xiaoting Lv, Changsheng
  Ma, Manal Alshehri, Inji Jaber, Faroug Tifratene, Wei Wang, Takashi Gojobori,
  Carlos~M. Duarte, Xin Gao, and Xiangliang Zhang.
\newblock {SenWave}: Monitoring the global sentiments under the covid-19
  pandemic.
\newblock {\em arXiv preprint arXiv:2006.10842}, 2020.

\end{thebibliography}
\end{document}